\documentclass[10pt,twocolumn,letterpaper]{article}

\usepackage{cvpr}              

\usepackage{graphicx}
\usepackage{amsmath}
\usepackage{amssymb}
\usepackage{booktabs}

\usepackage[pagebackref,breaklinks,colorlinks]{hyperref}

\usepackage[capitalize]{cleveref}
\crefname{section}{Sec.}{Secs.}
\Crefname{section}{Section}{Sections}
\Crefname{table}{Table}{Tables}
\crefname{table}{Tab.}{Tabs.}

\def\cvprPaperID{6} 
\def\confName{CVPR}
\def\confYear{2025}

\begin{document}

\title{BodyGPS: Anatomical Positioning System}

\author{
Halid Ziya Yerebakan \and
Kritika Iyer \and
Xueqi Guo \and
Yoshihisa Shinagawa \and
Gerardo Hermosillo Valadez \\
Siemens Healthineers \\
40 Liberty BLVD, Malvern, PA \\
{\tt\small firstname.lastname@siemens-healthineers.com}
}

\maketitle
\begin{abstract}

We introduce a new type of foundational model for parsing human anatomy in medical images that works for different modalities. It supports supervised or unsupervised training and can perform matching, registration, classification, or segmentation with or without user interaction. We achieve this by training a neural network estimator that maps query locations to atlas coordinates via regression. Efficiency is improved by sparsely sampling the input, enabling response times of less than 1 ms without additional accelerator hardware. We demonstrate the utility of the algorithm in both CT and MRI modalities.

\end{abstract}

\section{Introduction}
\label{sec:intro}

Various forms of anatomical intelligence are currently in use in the medical imaging industry and research, such as registration, segmentation, landmarking, and bounding box detection for clinical automation tasks such as longitudinal comparison, surgery planning, or scanner automation. Each approach has its own advantages and limitations. However, central to all of these methods is the task of estimating the semantic anatomical location of any given point. In this paper, we introduce a new formulation of anatomical intelligence via a foundational model using efficient deep learning regression methods. Although anatomical size and shape vary among individuals, the human body is generally consistent. This consistency allows us to define a coordinate system for semantic locations, similar to global geographical reference systems such as GPS. We refer to our method and task as BodyGPS. Our method creates new opportunities for medical AI, allowing different use cases without additional task-specific training.

Although similar objectives can be achieved by registering images in a common atlas and aligning individual scans to a standardized anatomical template, such approaches are computationally intensive and require carrying the atlas image for all applications. Additionally, they do not utilize existing datasets to learn variations of the corresponding body parts. In contrast, our proposed BodyGPS system employs a small, learned neural network to estimate semantic anatomical locations directly, without requiring image registration to an atlas. This results in a more flexible and scalable solution for mapping anatomical coordinates, especially when only a few points need to be mapped.

By estimating semantic positions, our system provides navigation functionality for the human body. An unlimited number of landmark positions can be queried by requesting the displacement vector to a target location. In other words, a landmark detector can be derived from a single-point prompt from the original atlas, effectively enabling one-shot learning (even zero-shot since image data in the prompt are only for visualization for the user).

We present the details of the efficient regression method for BodyGPS in Section 3. In Section 4, we demonstrate the effectiveness of our approach in the Total Segmentator dataset through a series of experiments, comparing our results with existing methods. Finally, we propose directions for future research and applications.

\begin{figure*}
    \centering
    \includegraphics[width=\linewidth]{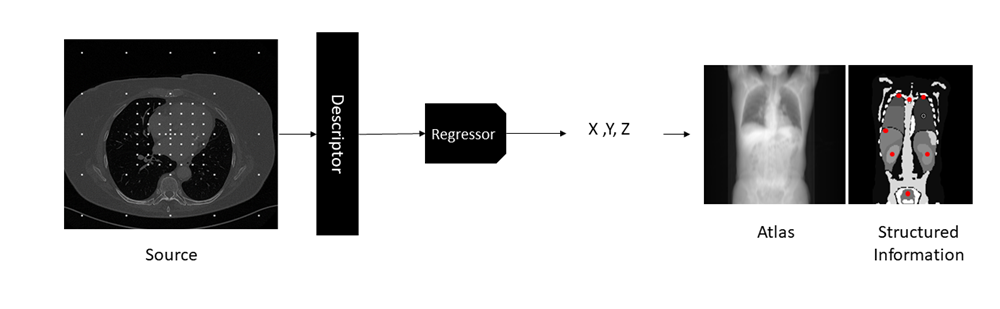}
    \caption{BodyGPS is a deep learning regressor that maps any query point to reference atlas coordinates}
    \label{fig:bodygps}
\end{figure*}

\section{Related Work}
Recent medical imaging research has focused primarily on the experimentation with diverse neural network architectures tailored to specialized tasks, often at the expense of exploring alternative problem formulations. However, in many practical scenarios, problem formulation and data quality influence performance more significantly than network architecture choices. Nevertheless, there has been continuous progress in developing improved and efficient models for various medical imaging tasks.

A notable milestone in this domain is the U-Net architecture, widely adopted for medical image segmentation due to its superior performance over prior approaches \cite{ronneberger2015u}. Its encoder-decoder structure with skip connections effectively captures both high-level context and fine-grained details, making it highly successful across multiple segmentation applications on many modalities. U-Net's success has inspired new hierarchical architectural designs \cite{zhou2019unet++}. More recently, transformer-based models have further advanced segmentation accuracy, as evidenced by improved Dice scores \cite{shaker2024unetr++}. Numerous subsequent studies have adapted similar architectures for diverse medical imaging tasks \cite{balakrishnan2019voxelmorph,noothout2020deep,yan2018deeplesion,rood2019toward,iuga2021automated}.

Within the last 5 years, self-supervised learning and foundational models have gained attention in medical imaging. For example, Vox2vec \cite{goncharov2023vox2vec} employs contrastive learning for multi-scale voxel-level segmentation, and MedImageInsight \cite{codella2024medimageinsight} is optimized for classification and image retrieval tasks. Additionally, \cite{yao2021one} and UAE \cite{bai2025uae} targeted learning embeddings with multiple scales on medical images for landmark detection and matching. However, embeddings in these models lack physical interpretability. In a parallel direction, despite being a non-learning method, the point matching approach presented by Yerebakan et al. \cite{pointmatching} demonstrates foundational capabilities that enable prompt-based classification and landmarking.

Although regression methods have been proposed, they remain less prevalent in the literature. Lei et al. \cite{lei2021contrastive} addressed relative displacement regression using iterative refinement dependent on reference images. Our approach introduces a novel foundational model capable of adapting to various tasks by directly producing absolute, interpretable coordinates while eliminating run-time dependencies with high efficiency.

\section{Method}

BodyGPS method consists of two steps: first, efficient sparse descriptor curation for the query point, and second, a regression algorithm based on a deep learning model.

\subsection{Sparse Descriptor Extraction}

Descriptors are generated through sparse sampling of intensity values based on predefined offset locations in a grid specified in millimeter displacements at different resolutions for any query point. Instead of resampling images, the sampler function is adjusted by voxel spacings in the image, thereby avoiding extra memory and computational overhead. An illustrative example of the location of the sparse sampling intensity is shown in Figure \ref{fig:bodygps}. For any query point, the descriptor is computed by performing a memory lookup at determined offset positions and assembling these into a 1D array. Thus, almost no computation is necessary.

In our experiments, we specifically employed combined 2D and 3D grids to sample intensities for the creation of descriptors. The 2D component consists of three orthogonal planes, each defined by a 27x27 grid at a 4 mm resolution. The 3D component comprises seven three-dimensional grids at resolutions of 2, 3, 5, 8, 12, 28, and 64 mm, respectively, each using a 9x9x9 grid to accommodate varying slice thicknesses commonly found in CT images. Spacing increments use non-integer multipliers in general to prevent redundant sampling at the same locations across different resolutions. The grid size of 9 allows for simultaneous visualization of 2D and 3D sampling, facilitating debugging as shown in the Sudoku pattern \cite{yerebakan2024real}. We named this method "3.5D sampling" because it incorporates more than purely 3D information.

\subsection{Mapping to Atlas}

An atlas is a reference volume. Annotating a single atlas image is simpler compared to machine learning dataset curation since it is a one-time effort. Also, locations of points on the atlas image can be easily associated with semantic anatomical information. Estimating the mapping of any point from the source image to the atlas image associates this anatomical information with the point of interest in the source image, as illustrated in Figure \ref{fig:bodygps}.

For example, if the question is about identifying the organ at a selected point, the segmentation mask of the atlas image can instantly provide the organ label once the mapping is complete. Another example involves determining the distance from a point of interest to a specific anatomical landmark. In this case, we can quickly reference the landmark location in the atlas image and find the location by giving the same normalized coordinates to the image of interest as a navigation system.

We have used a deep neural network regression method to create a mapping to the atlas coordinates. The input to the neural network regressor is sparse descriptors for a given query point. The output is the normalized coordinates according to the atlas, such as centering the carina at location (0,0,0). While the regressor network could be any machine learning regression model, in our specific case, we used a 16-layer residual network that processes 240-dimensional flat vectors after input projections. Finally, in the last layer, a three-dimensional linear function outputs normalized x, y, z coordinates.

The primary challenge in training this method is to establish ground truth for normalized coordinates. Manual annotation of mapping points as supervision is feasible for only a few landmarks. Therefore, automated registration algorithms can be utilized to create mappings from input images to atlas coordinates. The slow speed of non-rigid registrations is acceptable here, as it only incurs a minor one-time training cost.
At runtime, the trained model takes input from the selected point and estimates the normalized coordinate. This normalized coordinate can be mapped back to the actual atlas location by adding the carina offset. If necessary, multiple queries in nearby regions could be executed to enhance robustness.

\section{Experiments}

We have evaluated the proposed approach for segmentation and matching tasks on CT. Additionally, we have demonstrated a supervised version on landmark detection of MRI ankle images.

\subsection{Segmentation}

In our CT experiments we have used Total Segmentator dataset \cite{wasserthal2023totalsegmentator}. The case "s1045" is selected as an atlas due to larger field of view. All other images are registered to the atlas using Rigid + Non-rigid registration \cite{hermosillo2002variational}. Then 1500 points were randomly selected from each image and additional 1500 were added within a neighborhood of those points by small perturbation.  We have utilized logMSE loss and trained for 1000 epochs. No supervision is used except for the mask of the atlas. 


The segmentation task is to obtain the organ labels corresponding to the voxels. We used a 3mm grid to query BodyGPS coordinate estimations. Once the coordinates are known, the atlas mask is used to determine the label. Transferring all labels back to the source image will construct a segmentation map. In an interactive mode, one could query mouse location in real time to get organ labels that will serve as a classifier.

\begin{figure*}
    \centering
    \begin{subfigure}[b]{0.25\textwidth}
        \centering
        \includegraphics[width=\textwidth]{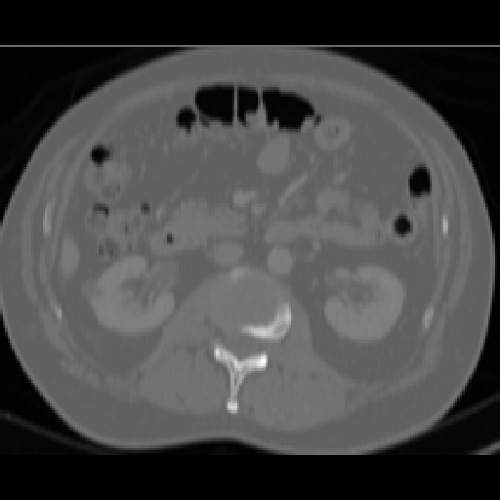}
        \caption{Input Image}
        \label{fig:sub1}
    \end{subfigure}
    \hfill
    \begin{subfigure}[b]{0.25\textwidth}
        \centering
        \includegraphics[width=\textwidth]{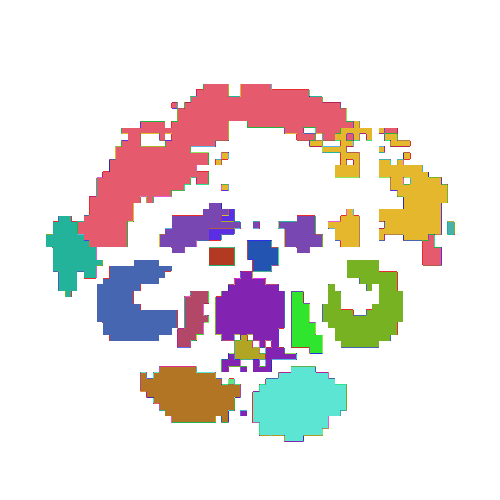}
        \caption{BodyGPS Estimate}
        \label{fig:sub2}
    \end{subfigure}
    \hfill
    \begin{subfigure}[b]{0.25\textwidth}
        \centering
        \includegraphics[width=\textwidth]{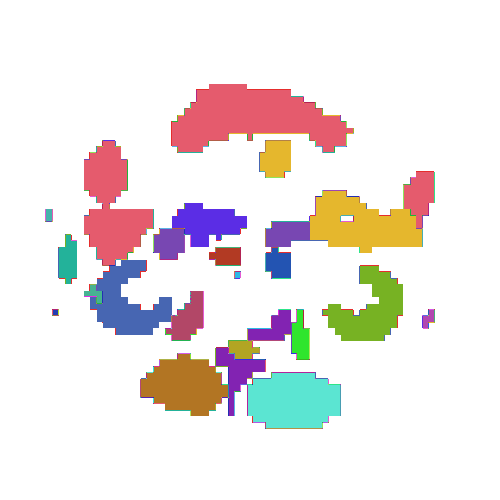}
        \caption{Ground Truth Mask}
        \label{fig:sub3}
    \end{subfigure}
    \caption{Application of BodyGPS on segmentation}
    \label{fig:segmentation}
\end{figure*}

The computational time for full volume is slow, on the order of minutes. However, any grid sampling could be used to change the trade-off between speed and accuracy without changing the model. A single query takes less than 1ms. A qualitative example is demonstrated in Figure \ref{fig:segmentation}.

We report the micro-average Dice scores for all organs present in the test images. For comparison, we used published foundational models trained with mask supervision. Despite being unsupervised, our method outperforms SAM-Med3D and FastSAM3D, although it performs worse than the SAMU method, as shown in Table \ref{segmentation}.

\begin{table}[]
\centering
\caption{Dice Scores of Segmentation Results}\label{tab1}

\begin{tabular}{l|l|}
\cline{2-2}
                                                                          & Total Segmentator \\ \hline
\multicolumn{1}{|l|}{SAM-Med3D \cite{sammed3d}}           & 0.334             \\ \cline{1-2}
\multicolumn{1}{|l|}{FASTSAM3D \cite{shen2024fastsam3d}}  & 0.242              \\ \cline{1-2}
\multicolumn{1}{|l|}{SAMU \cite{bae2024samu}}             & 0.756   \\ \hline
\multicolumn{1}{|l|}{BodyGPS}                             &  0.664   \\ \hline
\end{tabular}
\label{segmentation}
\end{table}

\subsection{Matching}

We utilized the same self-supervised pretrained BodyGPS network to evaluate the matching task between longitudinal studies in an in-house CT dataset with 348 pairs. We compared BodyGPS to PointMatching \cite{pointmatching} in this study.

\begin{figure}
    \centering
    \includegraphics[width=\linewidth]{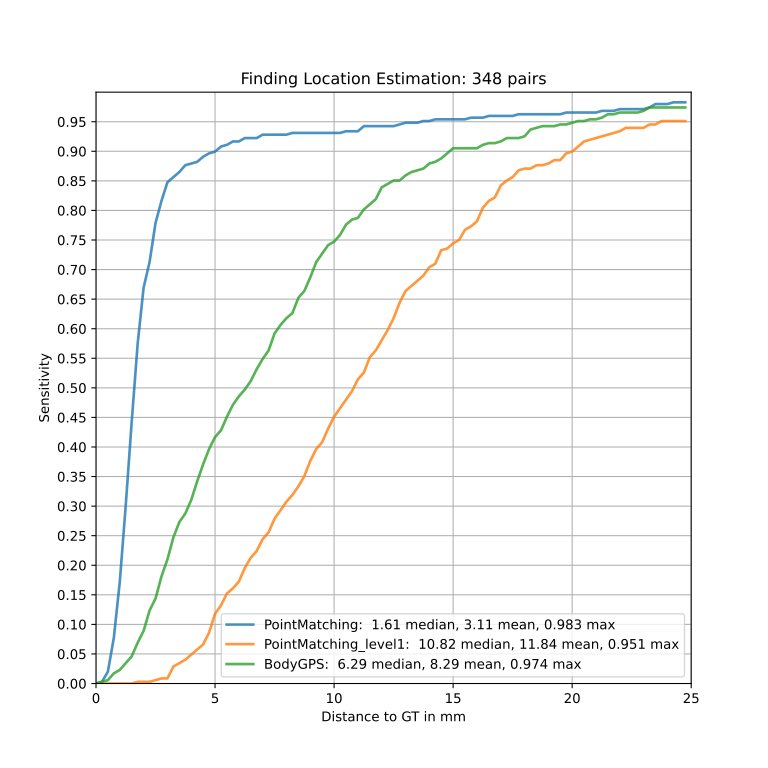}
    \caption{Performance of Methods For Longitudinal Matching}
    \label{fig:nuremberg}
\end{figure}

In the matching task, BodyGPS is used to normalize the query point in the source image, and the corresponding point is found with 50 navigation iterations in the target image. We evaluated methods with sensitivity at different distance thresholds. Our results are shown in Figure \ref{fig:nuremberg}. BodyGPS performs better than PointMatching in the first level. However, in subsequent levels, point matching generates more precise results due to utilization of pixel information at multiple scales. We can consider coarse-to-fine strategies for BodyGPS in future studies.

\subsection{Landmarking}

\begin{figure}
    \centering
    \includegraphics[width=\linewidth]{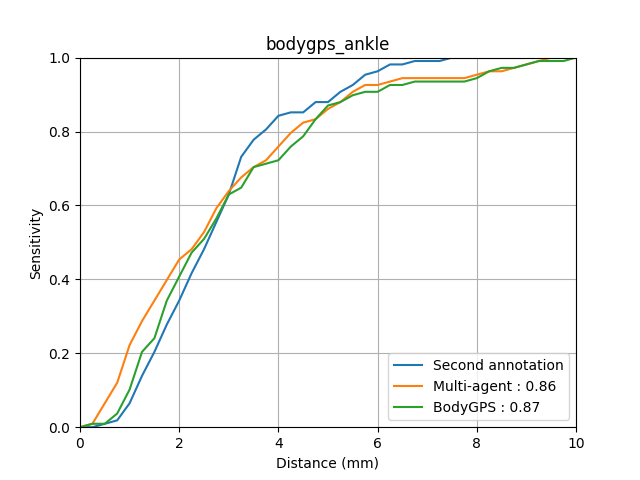}
    \caption{Landmark location estimation with supervised BodyGPS}
    \label{fig:fibula}
\end{figure}

In our last experiment, we have utilized the BodyGPS regression mechanism on a supervised fibula landmark dataset on MRI ankle images. 172 series for training and 109 series for testing were annotated for the landmark. We had a second annotation on the test set to check human performance. BodyGPS is used to estimate displacement to the landmark in world coordinates and iterated multiple times to find the target landmark like a navigation system. We used center of the image as starting point. The FROC curve of the results is shown in Figure \ref{fig:fibula} with sensitivity at 5mm in the legend. It can be seen that BodyGPS is more precise than the second annotation in the precision region and reaches 100\% at less than 10mm. Additionally, a multi-agent version of the navigation with different start points demonstrates further improved precision. 

\section{Discussions}

Our approach relies on unsupervised registration methods to generate ground truth, with registration quality directly affecting final performance. Notably, Jena et al. \cite{jena2024deep} demonstrate that traditional registration methods outperform deep-learning-based approaches in multimodal tasks, raising questions about the actual effectiveness of deep-learning registration. This prompts us to examine whether point matching (i.e. prompt-based registration) might outperform BodyGPS in certain contexts. Nevertheless, BodyGPS offers substantial speed advantages, which could enable hybrid strategies combining both methods’ strengths. Additionally, a hierarchical coarse-to-fine scheme could further enhance estimation precision. Beyond our current applications, this framework maps different radiological modalities into a unified anatomical space—facilitating cross-modality alignment, retrieval, and abnormality detection via comparison with normal anatomy atlases. Our solution thus provides interoperability across medical systems.

\section{Conclusion}

We have demonstrated a new foundational model that generates a semantic coordinate system for medical image queries using 3.5D sparse sampling and a 1D residual network. Our experiments validate the effectiveness of this coordinate estimation for segmentation, classification, and landmarking. The BodyGPS algorithm opens new opportunities for more efficient parsing of human anatomy, achieving query response times under 1 ms. Future work will focus on exploring hierarchical coarse-to-fine strategies and cross-modality applications to further enhance the precision and versatility of the framework.



{\small
\bibliographystyle{ieee_fullname}
\bibliography{bodygps}
}

\end{document}